\documentclass{article}

\usepackage{arxiv}

\usepackage[utf8]{inputenc}
\usepackage[T1]{fontenc}
\usepackage{hyperref}
\usepackage{url}
\usepackage{booktabs}
\usepackage{amsfonts}
\usepackage{amsmath,amssymb}
\usepackage{nicefrac}
\usepackage{microtype}
\usepackage{graphicx}
\usepackage{natbib}
\usepackage{doi}
\usepackage{algorithmic}
\usepackage{listings}
\usepackage{multirow}
\usepackage{xcolor}

\lstdefinelanguage{json}{
    basicstyle=\ttfamily\footnotesize,
    numbers=none,
    stepnumber=1,
    breaklines=true,
    showstringspaces=false,
    frame=single,
    columns=flexible,
    morestring=[b]",
    stringstyle=\color{black},
    morecomment=[l]{//},
    commentstyle=\color{gray}
}

\title{Can LLM-Reasoning Models Replace Classical Planning?\\ A Benchmark Study}

\author{%
  \href{https://orcid.org/0000-0001-5074-3652}{\includegraphics[scale=0.06]{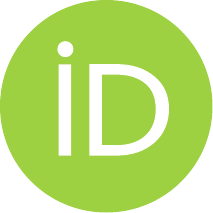}\hspace{1mm}Kai Goebel} \\
  Center for Vision, Automation \& Control\\
  AIT Austrian Institute of Technology GmbH\\
  Vienna, Austria \\
  \texttt{kai.goebel@ait.ac.at} \\
  \And
  \href{https://orcid.org/0000-0001-7846-1753}{\includegraphics[scale=0.06]{orcid.pdf}\hspace{1mm}Patrik Zips} \\
  Center for Vision, Automation \& Control\\
  AIT Austrian Institute of Technology GmbH\\
  Vienna, Austria \\
  \texttt{patrik.zips@ait.ac.at} \\
}

\renewcommand{\shorttitle}{LLMs vs Classical Planning for Robotics}

\hypersetup{
  pdftitle={Can LLM-Reasoning Models Replace Classical Planning? A Benchmark Study for Robotic Tasks},
  pdfauthor={Kai Goebel, Patrik Zips},
  pdfkeywords={Large Language Models, Planning, Robotics, PDDL, Benchmarking},
}

\begin{document}
\maketitle

\begin{abstract}
Recent advancements in Large Language Models have sparked interest in their potential for robotic task planning.
While these models demonstrate strong generative capabilities, their effectiveness in producing structured and executable plans remains uncertain.
This paper presents a systematic evaluation of a broad spectrum of current state of the art language models, each directly prompted using Planning Domain Definition Language domain and problem files, and compares their planning performance with the Fast Downward planner across a variety of benchmarks.
In addition to measuring success rates, we assess how faithfully the generated plans translate into sequences of actions that can actually be executed, identifying both strengths and limitations of using these models in this setting.
Our findings show that while the models perform well on simpler planning tasks, they continue to struggle with more complex scenarios that require precise resource management, consistent state tracking, and strict constraint compliance.
These results underscore fundamental challenges in applying language models to robotic planning in real world environments.
By outlining the gaps that emerge during execution, we aim to guide future research toward combined approaches that integrate language models with classical planners in order to enhance the reliability and scalability of planning in autonomous robotics.
\end{abstract}


\section{Introduction}

Task planning has long been a central pillar in robotic systems, enabling autonomous agents to deliberate about sequences of actions, allocate resources, and adapt to changing constraints.
Traditionally, symbolic planners such as Fast Downward \cite{helmert2006fast} have been the primary tools for generating provably correct solutions from well-defined domains using the Planning Domain Definition Language (PDDL) \cite{McDermott1998PDDLthePD}. 
However, their reliance on precisely modeled PDDL files often leads to scalability and representation challenges when confronted with real-world uncertainty.

Meanwhile, Large Language Models (LLMs) such as GPT-4, Claude, and Llama variants have demonstrated promising capabilities in commonsense reasoning, which can be further enhanced through strategies like chain-of-thought prompting \cite{wei2022chain}.
These reasoning abilities enable LLMs to generate structured multi-step outputs, making them useful for high-level planning and even machine control.
Recent studies show that LLMs can produce plans for simpler tasks, leveraging their extensive pretraining on textual corpora.
For instance, approaches like ReAct \cite{REACT2023} and LLM-Planner \cite{Hee2212} illustrate how LLMs can autonomously generate high-level action plans in open-domain contexts.

Despite these advancements, reliability and executability remain pressing concerns.
LLMs often lack robust environment grounding \cite{Gramopadhye12210, Submission2204} and are prone to generating incorrect or misleading outputs.
This issue is especially critical for robotics, where a single erroneous action can risk safety or cause task failure.
Moreover, there is growing recognition that domain constraints, partial observability, and concurrency require more than just fluent text generation.
As Kambhampati \cite{Kambhampati2402, Kambhampati2403} points out, pure LLM-based outputs require external validation or symbolic checks to ensure correctness in non-trivial planning scenarios.

\begin{figure}[t]
    \centering
    \includegraphics[width=0.75\linewidth]{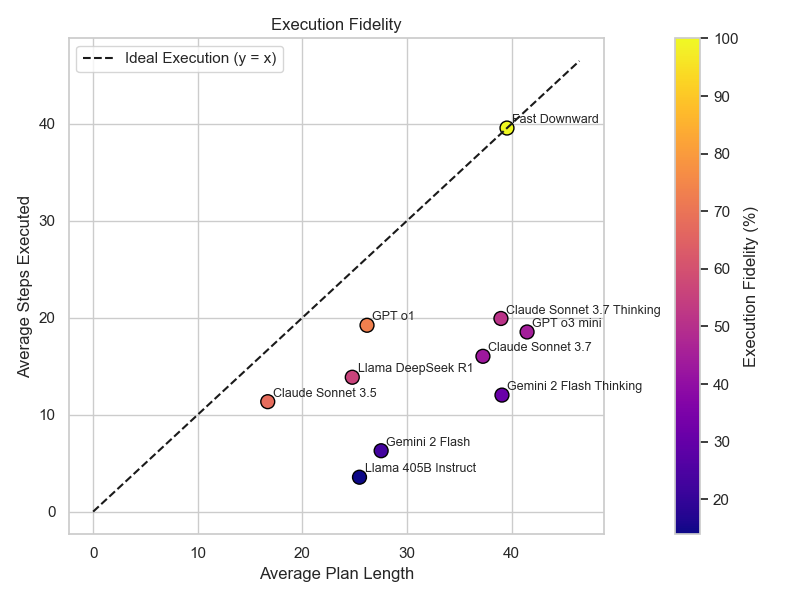}
    \caption{This scatter plot illustrates the relationship between average plan length (x-axis) and average executed steps (y-axis) for different LLM-based planners. The dashed diagonal ($y = x$) represents perfect execution fidelity, where all planned actions are successfully performed. Planners closer to this line exhibit better execution reliability. \textit{Fast Downward} aligns with the ideal execution line, indicating that all planned actions are executable. \textit{GPT-o1} has the highest execution fidelity (73.4\%), followed by \textit{Claude Sonnet 3.5} (67.9\%), both performing well in executing planned actions. In contrast, \textit{Llama DeepSeek R1} (55.9\%) and \textit{Claude Sonnet 3.7 Thinking} (51.1\%) demonstrate moderate execution fidelity, while \textit{Gemini 2 Flash} (22.8\%) and \textit{Llama 405B Instruct} (13.9\%) exhibit significant execution failures, indicating poor planning quality. One can also see that \textit{GPT-o3 mini} is the model that tends to overestimate the required plan lengths.}
    \label{fig:execution_fidelity}
\end{figure}

These converging trends, powerful but imperfect Large Language Model planners on one hand and proven classical planners on the other, have led to the development of hybrid solutions.
Several studies propose bridging the generative strengths of Large Language Models, particularly for domain knowledge and heuristics, with the search rigor of traditional planning tools \cite{Models2407} \cite{Liu2304}, often yielding better results in complex or partially observable tasks.
This synergy leverages the commonsense reasoning capabilities of Large Language Models while preserving the formal reliability of symbolic solvers, an arrangement that could be particularly transformative for real world robotics, where environment fluctuations and concurrency constraints are significant.

Building on this context, our study systematically compares the planning capabilities of nine LLMs and Fast Downward across five distinct PDDL domains: \textit{barman}, \textit{blocks}, \textit{elevator}, \textit{satellite}, and \textit{tidybot}.  
Each domain reflects varying degrees of complexity in resource management, concurrency, and object handling.  
We focus not only on raw success rates but also on execution fidelity, the extent to which generated plans translate into executable actions.  
Our results underscore the contrast between the guaranteed executability of symbolic planners in many domains and the strong yet inconsistent performance of LLMs as planners.  
While LLMs excel in simpler tasks, they struggle in domains that place heavy demands on resource coordination.  
These findings reinforce the notion that although LLM based approaches show considerable promise, they require reliable integration with classical planning frameworks to ensure consistent performance.

\begin{figure}[t]
    \centering
    \includegraphics[width=0.75\linewidth]{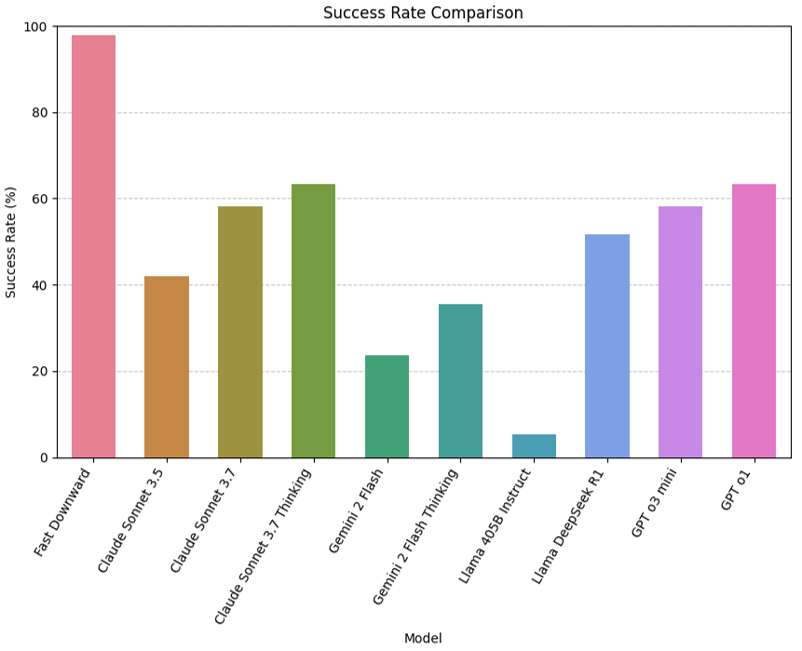}
    \caption{Comparison of the planning capabilities of nine different LLMs against the classical planner \textit{Fast Downward}, which serves as the baseline. This figure presents the average success rate of solving the given problems across multiple PDDL domains. Among the LLM-based planners, \textit{GPT-o1} and \textit{Claude Sonnet 3.7 Thinking} achieve the highest success rate, both at 63.4\%. Notably, the reasoning model \textit{Claude Sonnet 3.7 Thinking} outperforms its non-reasoning counterpart \textit{Claude Sonnet 3.7} (58.1\%), and \textit{Gemini 2 Flash Thinking} (35.5\%) shows an improvement over \textit{Gemini 2 Flash} (23.7\%). However, even the best-performing models do not achieve robust planning. Further analysis, particularly in more complex domains, shows that LLMs still struggle, raising concerns about whether their success in well-known domains stems from genuine reasoning or merely from recalling solutions encountered during extensive pretraining. These findings highlight the need for plan validation and execution fidelity in LLM-driven planning for robotic tasks.}
    \label{fig:success_rate}
\end{figure}
\begin{figure*}[t]
    \vspace{2mm}
    \centering
    \includegraphics[width=\textwidth]{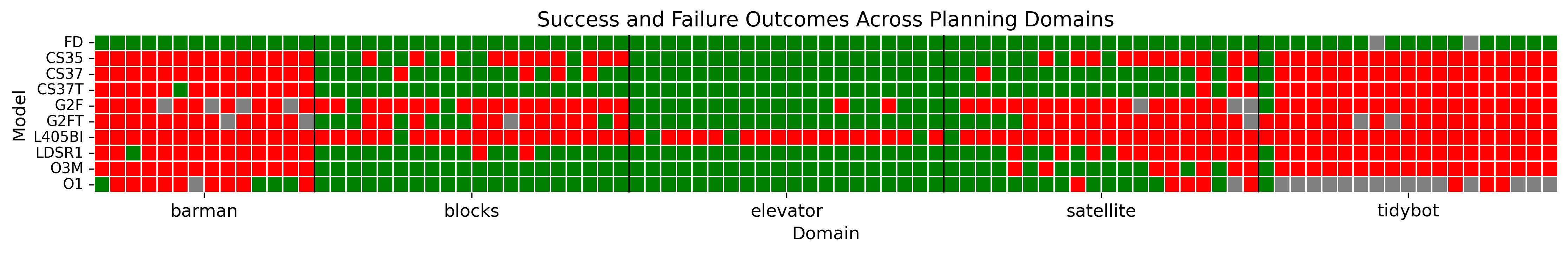}
    \caption{
        Per-domain success and failure outcomes of the evaluated LLMs compared to the baseline classical planner \textit{Fast Downward} (FD). The evaluated models include CS35: \textit{Claude Sonnet 3.5}, CS37: \textit{Claude Sonnet 3.7}, CS37T: \textit{Claude Sonnet 3.7 Thinking}, G2F: \textit{Gemini 2 Flash}, G2FT: \textit{Gemini 2 Flash Thinking}, L405BI: \textit{Llama 405B Instruct}, LDSR1: \textit{Llama DeepSeek R1}, O3M: \textit{GPT-o3 mini}, and O1: \textit{GPT-o1}. Success cases are visualized in green, failures in red, and cases where no plan was generated in gray. 
        The five evaluated domains, \textit{barman}, \textit{blocks}, \textit{elevator}, \textit{satellite}, and \textit{tidybot}, are each represented with their respective problem sets. In the \textit{tidybot} domain, \textit{Fast Downward} failed to find solutions for two problems. Overall, the best performing models are \textit{Claude Sonnet 3.7 Thinking} and \textit{GPT o1}. Notably, \textit{GPT o1} sometimes concluded that no solution existed after reasoning, leading to no plan generation. In contrast, the two \textit{Gemini} models failed to generate plans in some cases due to their limited maximum output tokens.
        Performance across domains varies: models perform well in the \textit{blocks} and \textit{elevator} domains, followed by the \textit{satellite} domain, while all struggle with the \textit{barman} and \textit{tidybot} domains. It remains unclear whether success in certain domains results from genuine reasoning or if models are recalling solutions from their pretraining data. The high success rate of non-reasoning models in the \textit{elevator} domain suggests that this domain may be covered in the pretraining corpus. In the \textit{blocks} domain, improvements from newer reasoning models are evident, while the \textit{barman} and \textit{tidybot} domains remain challenging. 
    }
    \label{fig:success_timeline}
\end{figure*}
\section{Related Work}
\label{sec:related_work}

Task planning has traditionally relied on symbolic methods, where planners generate optimal action sequences from well-defined representations, such as those defined in the Planning Domain Definition Language (PDDL). 
While methods like Fast Downward offer efficiency and formal verification, they struggle with uncertainty and require significant domain engineering. 
To mitigate these limitations, recent approaches explore LLM-based planning and hybrid methods that integrate symbolic reasoning with generative capabilities.

\subsection{Emergence of LLM-Based Planning}
\subsubsection{Direct Use of LLMs as Planners}
A growing body of research explores directly leveraging LLMs to generate multi step plans without explicit symbolic backends.  
In this approach, the LLM takes a task description as input and outputs an action sequence, often in natural language, for execution by an agent or robot.  
For instance, Huang \emph{et al.}~\cite{Huang2201} demonstrate ``zero shot'' action planning for embodied agents.  
Similarly, ReAct~\cite{REACT2023} interleaves reasoning traces and actions within a single inference loop.

Though promising, purely LLM-driven planning has limitations.
Chen \emph{et al.}~\cite{Chen2408} show that LLMs often fail to incorporate critical constraints in long-horizon tasks, while Cloos \emph{et al.}~\cite{Cloos2407} highlight that LLMs frequently overlook basic logical rules in puzzle-like environments.
To mitigate these issues, Hee \emph{et al.}~\cite{Hee2212} introduce LLM-Planner, embedding environment grounding and few-shot examples into the prompt.
Recent frameworks, such as AgentGen~\cite{Generation2408}, further emphasize environment and task generation to enhance the diversity of LLM-based planning.

Overall, LLMs' ability to avoid hallucinated or contradictory steps remains a challenge.
As shown by Kambhampati~\cite{Kambhampati2402} and Valmeekam~\cite{Valmeekam2305}, purely autoregressive outputs often lack robust self-verification mechanisms.

\subsubsection{Hybrid PDDL + LLM Approaches}
Another strand of research fuses PDDL-based planners with LLMs, aiming to leverage the strengths of both.
Symbolic planners provide rigorous state-space search and correctness guarantees, while LLMs contribute commonsense reasoning and semantic understanding.
For instance, Capitanelli and Mastrogiovanni~\cite{Large2303} propose Teriyaki, a neurosymbolic planner where \textit{GPT-3} generates PDDL plans.
Stein \emph{et al.}~\cite{Stein2311} automate prompt generation from PDDL domains, while Silver \emph{et al.}~\cite{Silver2305} show that \textit{GPT-4} can internalize domain constraints to produce scalable plan generators in code form.

Further techniques emphasize iterative refinement and environment feedback.
Mahdavi \emph{et al.}~\cite{Models2407} create feedback loops between an LLM-generated PDDL file and a running simulator.
Zhou \emph{et al.}~\cite{Zhou2308} propose iterative self-refinement~(ISR-LLM), incorporating a validator to revise generated plans.
Other approaches strive to balance efficiency and accuracy. For instance, Kwon \emph{et al.}~\cite{Kwon2409} utilize neuro-symbolic goal decomposition to achieve this balance.

\subsubsection{LLM-Based Task Planning in Robotics}
LLMs are increasingly applied in robotics, where planning needs to consider physical constraints, sensor noise, and complex dynamics.
Several studies demonstrate how LLMs can assist in or autonomously devise robotic action plans.
Capitanelli and Mastrogiovanni~\cite{Large2303} extend GPT-3 with Teriyaki, while Joublin \emph{et al.}~\cite{Joublin2310} propose CoPAL, an iterative feedback method for real-time correction of robot action plans.
Efforts such as SayCan~\cite{Raedt2308} ground LLM-based reasoning with learned value functions, whereas PaLM-E~\cite{Submission2303} integrates sensor input with textual prompts for multimodal embodied planning.

\subsection{Benchmarks and Evaluation Metrics for LLM Planning}
\subsubsection{Classical Benchmarks (PDDL-Focused)}
Efforts to systematically evaluate LLM-based planners often rely on classical IPC-style domains.
PlanBench~\cite{Change2206} assesses LLMs on PDDL problems, highlighting common errors.
A follow-up study by Valmeekam \emph{et al.}~\cite{Valmeekam2409} contrasts standard LLMs with ``large reasoning models.''
The Planetarium framework~\cite{Zuo2407} introduces automated checks for semantic equivalence in PDDL, while frameworks such as NL2Plan~\cite{Gestrin2405} investigate text-to-PDDL conversion.

\subsubsection{Everyday Tasks and Natural Language Benchmarks}
Another research direction focuses on natural-language planning scenarios.
NATURAL PLAN~\cite{Steven2406} evaluates LLMs on multi-constraint tasks, while TravelPlanner~\cite{Chen2408} challenges LLMs with travel plans.
Meanwhile, ``Baba Is AI''~\cite{Cloos2407} is a puzzle-like environment designed to test LLM reasoning.

\subsubsection{Emerging Metrics}
Traditional planning metrics include success rate, plan length, and runtime. 
However, execution fidelity, defined as the extent to which generated plans translate into executable actions, is also an important consideration. 
Open Grounded Planning~\cite{Guo2406} tests text-based plans, while CAT-BENCH~\cite{Lal2406} evaluates causal-temporal ordering.

\section{Methods}

This study examines the planning capabilities of various LLMs compared to Fast Downward, a well-established classical planner for task planning, using PDDL domain and problem descriptions.
This section details the evaluation setup, the LLMs assessed, the domains selected for benchmarking, the prompts employed to generate plans, and the evaluation metrics utilized.

\subsection{Fast Downward Baseline}
Fast Downward is a domain-independent planning system that translates PDDL tasks into multi-valued variable representations, highlighting implicit constraints and dependencies. 
We use the \texttt{seq-sat-lama-2011} heuristic configuration for our experiments. This configuration integrates landmark and FF heuristics within an iterated search scheme, balancing speed and plan quality through a sequence of lazy greedy best-first searches and weighted \texttt{A*} steps. 
This anytime approach refines initial solutions, allowing the planner to improve plan quality if time remains. 
In our setup, we cap the planning time at 600 seconds, offering a robust classical benchmark against which LLM-based methods can be compared. 
By leveraging hierarchical decomposition and heuristic guidance, Fast Downward efficiently handles diverse domains, reinforcing its position as a reliable point of reference for evaluating emerging LLM-based planners.

\subsection{Overview of evaluated LLMs}

In our study, we evaluate the LLMs listed in Table \ref{tab:llms_overview}.
Given the rapid advancements in the field, this evaluation represents a snapshot of the current state of the art.
We acknowledge the release of new models such as \textit{Grok-3} and \textit{DeepSeek-R1}, which were not included in our evaluation due to inaccessibility at the time of writing.
To include the reasoning-optimized \textit{DeepSeek-R1} model in our study, we used the \textit{Llama DeepSeek R1} version, which is a distilled version of the original model.
We accessed the models \textit{Llama DeepSeek R1} and \textit{Llama 405B Instruct} via the provider \texttt{fireworks.ai} \cite{FireworksAI2024}, a platform that hosts and provides access to a variety of AI models.

\begin{table*}[t]
    \vspace{4mm}
    \centering
    \renewcommand{\arraystretch}{1.1}
    \caption{Evaluated LLMs}
    \label{tab:llms_overview}
    \resizebox{\textwidth}{!}{%
    \begin{tabular}{p{3.3cm} p{1.9cm} p{7.0cm} p{2cm} p{1.2cm}} 
        \toprule
        \textbf{Model} & \textbf{Type} & \textbf{Reasoning Capabilities} & \textbf{Release Date} & \textbf{GPQA} \\
        \midrule
        Claude Sonnet 3.5 & Proprietary & 
        \raggedright Advanced multi-step reasoning and code generation, demonstrating superior logical inference. & 
        June 2024 & 
        59.4 \\
        
        Claude Sonnet 3.7 & Proprietary & 
        \raggedright Hybrid reasoning model with improved structured problem-solving. & 
        Feb 2025 & 
        68.0 \\

        Claude Sonnet 3.7 Thinking & Proprietary & 
        \raggedright Extended thinking mode enabling detailed, step-by-step reasoning for complex problem-solving. & 
        Feb 2025 & 
        84.8 \\
        
        Gemini 2 Flash & Proprietary & 
        \raggedright General-purpose multimodal LLM optimized for efficiency but lacks explicit reasoning optimization. & 
        Dec 2024 & 
        58.6 \\
        
        Gemini 2 Flash Thinking & Proprietary & 
        \raggedright Experimental variant specifically designed for enhanced reasoning and problem-solving tasks. & 
        Jan 2025 & 
        74.2 \\
        
        Llama 405B Instruct & Open-Source & 
        \raggedright State-of-the-art open-source model with competitive reasoning performance. & 
        July 2024 & 
        50.7 \\
        
        Llama DeepSeek R1 & Open-Source & 
        \raggedright Optimized for reasoning via reinforcement learning with robust chain-of-thought solutions. & 
        Jan 2025 & 
        71.5 \\
        
        GPT-o3 mini & Proprietary & 
        \raggedright Compact reasoning-optimized model trained for enhanced problem-solving. & 
        Jan 2025 & 
        79.7 \\
        
        GPT-o1 & Proprietary & 
        \raggedright Explicit chain-of-thought reasoning strategies. & 
        Dec 2024 & 
        75.7 \\
        \bottomrule
    \end{tabular}
    }
\end{table*}

To provide a reference point for the reasoning capabilities of the evaluated LLMs, Table \ref{tab:llms_overview} includes a brief description of each model's reasoning capabilities and their GPQA Diamond Accuracy \cite{rein2023gpqagraduatelevelgoogleproofqa}.
The GPQA Diamond benchmark consists of 448 challenging multiple-choice questions in biology, physics, and chemistry, designed to be difficult for both human experts and AI models.
Higher accuracy on the GPQA Diamond benchmark indicates superior reasoning capabilities.
The accuracy values in the table are collected from various sources \cite{DeepSeekVsOpenAI2025, MetaReleasesLlama3.1-405B2025, Olteanu2025, Sharma2025, xAI2025, Anthropic2025}.

\subsection{Domains Evaluated}
These benchmarking domains cover a range of manipulation, scheduling, and resource-management challenges. Below, we summarize each domain’s main features:

\begin{itemize}
    \item \textbf{Barman (IPC 2014, 14 problems)}: Involves a robot bartender preparing and serving drinks using tools like dispensers and shakers. Each glass must be empty and clean before filling, and only one object can be held at a time, introducing tight sequencing constraints.
    
    \item \textbf{Blocks (IPC 2000, 20 problems)}: A classic typed STRIPS domain where a robot arm stacks blocks into a goal configuration. It tests careful action ordering, ensuring a block is clear before it can be moved, revealing combinatorial challenges.
    
    \item \textbf{Elevator (IPC 2000, 20 problems)}: Simulates transporting passengers across multiple floors. Planners must handle embarkation and disembarkation, optimizing routes to deliver passengers efficiently. Proper sequencing is required to avoid missed pickups and minimize travel time.
    
    \item \textbf{Satellite (IPC 2004, 20 problems)}: Focuses on coordinating multiple satellites to collect image data. Satellites must be calibrated and oriented correctly to capture targets. Scheduling transmissions and adjusting instruments within time windows adds resource-management complexity.
    
    \item \textbf{Tidybot (IPC 2011, 19 problems)}: Models a household cleaning task. One or more robots navigate a 2D grid, picking up items and placing them onto goal locations like tables or cupboards. Gripper reach is limited, and objects can obstruct each other, often requiring the use of carts to move multiple items efficiently.
\end{itemize}

\subsection{LLM Prompt for Plan Generation}
To ensure consistency and interpretability in LLM-generated plans, we design our prompt to encourage structured reasoning. The LLM first provides a high-level reasoning overview, outlining its strategy for solving the planning problem, followed by a step-by-step sequence of actions, each with an explicit justification.

\subsubsection{Task Description}
The LLM acts as a planning assistant that reads PDDL domain and PDDL problem descriptions, then generates a structured plan. The response must include:
\begin{itemize}
  \item A high-level reasoning section describing the overall approach.
  \item A detailed plan, listed as a sequence of actions.
\end{itemize}

\subsubsection{Input}
\begin{itemize}
  \item \textbf{PDDL domain:} Defines predicates, actions, and constraints.
  \item \textbf{PDDL problem:} Specifies objects, the initial state, and the goal conditions.
\end{itemize}

\subsubsection{Output Format}
The response must be valid JSON with two key components:
\begin{itemize}
  \item \textbf{Reasoning:} A high-level explanation describing the approach.
  \item \textbf{Plan:} An ordered list of actions. Each action object includes:
  \begin{itemize}
    \item \textit{name}: Identifier of the action (matching the domain).
    \item \textit{parameters}: Objects the action applies to.
    \item \textit{reason}: Justification for choosing this action.
    \item \textit{confirm\_reasoning}: A validation statement confirming correctness.
  \end{itemize}
\end{itemize}

Below the template of the JSON response format:

\vspace{3pt}

\lstset{
  language=json,
  basicstyle=\ttfamily\footnotesize,
  columns=flexible,
  breaklines=true,
  frame=single,
  showstringspaces=false
}

\begin{lstlisting}[linewidth=\columnwidth, xleftmargin=1.5mm, xrightmargin=1.5mm]
{
  "reasoning": [
    "A high-level explanation of the overall plan, describing how it transitions from the initial state to the goal."
  ],
  "plan": [
    {
      "name": "action_name",
      "parameters": ["arg1", "arg2"],
      "reason": "Explanation of why this action was chosen.",
      "confirm_reasoning": "Final validation statement."
    }
  ]
}
\end{lstlisting}

\subsection{Evaluation Metrics}

The following metrics were employed to assess the planners:

\begin{itemize}
    \item \textbf{Success Rate (SR)}: The fraction of problem instances successfully solved. This metric is essential for comparing the baseline capability of each planner. A solution is marked successful if the planner produces a valid plan that satisfies the domain’s goals.
    \item \textbf{Plan Length (PL)}: Measures the number of actions (or steps) in a generated plan. Classical planners typically produce near-optimal plans or follow heuristic-driven strategies, whereas LLM-based plans tend to be more variable. A plan that is too short may omit necessary actions, leading to incomplete execution, while excessively long plans increase the likelihood of errors or execution failures.
    \item \textbf{Executed Actions (Ac)}: Indicates how many actions from a generated plan can be validly executed according to the domain’s constraints. This highlights gaps between theoretical output and practical feasibility. LLM-based methods can struggle when the chain of actions is incomplete or violates preconditions.
    \item \textbf{Execution Fidelity}: The ratio of executed actions to planned actions, indicating how well the planner’s output aligns with the actual execution. A high fidelity score suggests that the planner generates plans that are more likely to be executable in practice.
    \item \textbf{Planning Time}: The time taken to generate a plan. This metric is crucial for real-time applications, where rapid decision-making is essential. LLM-based planners may require more time to reason through complex problems, potentially limiting their applicability in time-sensitive scenarios.
\end{itemize}

\begin{table}[t]
    \centering
    \caption{Average Planning Time per Model}
    \label{tab:planning_time}
    \begin{tabular}{l c}
        \toprule
        \textbf{LLM Model} & \textbf{Avg. Planning Time (s)} \\
        \midrule
        Claude Sonnet 3.5 & 14.22 \\
        Claude Sonnet 3.7 & 28.90 \\
        Claude Sonnet 3.7 Thinking & 112.61 \\
        Gemini 2 Flash & 15.13 \\
        Gemini 2 Flash Thinking & 22.02 \\
        GPT-o1 & 140.51 \\
        GPT-o3 mini & 98.61 \\
        Llama 405B Instruct & 27.08 \\
        Llama DeepSeek R1 & 160.15 \\
        \bottomrule
    \end{tabular}
\end{table}

\section{Results and Analysis}

\subsection{Success Rate \& Overall Planner Performance}

\begin{table*}[t]
    \vspace{3mm}
    \centering
    \caption{Performance Metrics for Planning Systems Across Five Domains. SR: Success Rate (\%), PL: Plan Length, Ac: Executed Actions}
    \label{tab:results_per_domain}
    \renewcommand{\arraystretch}{1.1} 
    \setlength{\tabcolsep}{3pt} 
    \footnotesize
    \resizebox{\textwidth}{!}{%
    \begin{tabular}{l rrr rrr rrr rrr rrr rrr}
    \hline
    \multirow{2}{*}{\textbf{Model}} 
    & \multicolumn{3}{c}{\textbf{Barman}} 
    & \multicolumn{3}{c}{\textbf{Blocks}} 
    & \multicolumn{3}{c}{\textbf{Elevator}} 
    & \multicolumn{3}{c}{\textbf{Satellite}} 
    & \multicolumn{3}{c}{\textbf{Tidybot}}
    & \multicolumn{3}{c}{\textbf{MEAN}} \\
    \cline{2-4} \cline{5-7} \cline{8-10} \cline{11-13} \cline{14-16} \cline{17-19}
     & \textbf{SR} & \textbf{PL} & \textbf{Ac} 
     & \textbf{SR} & \textbf{PL} & \textbf{Ac} 
     & \textbf{SR} & \textbf{PL} & \textbf{Ac} 
     & \textbf{SR} & \textbf{PL} & \textbf{Ac} 
     & \textbf{SR} & \textbf{PL} & \textbf{Ac} 
     & \textbf{SR} & \textbf{PL} & \textbf{Ac} \\
    \hline
    Fast Downward       & 100.0 & 98.7 & 98.7 & 100.0 & 26.9 & 26.9 & 100.0 & 10.4 & 10.4 & 100.0 & 40.0 & 40.0 & 89.5 & 39.5 & 39.5 & 97.85 & 39.56 & 39.56 \\
    Claude Sonnet 3.5   &   0.0 & 14.6 &  3.9 &  45.0 & 19.6 & 14.1 & \textbf{100.0} & 10.3 & 10.3 &  45.0 & 23.6 & 23.1 &  \textbf{5.3} & 14.8 &  2.7 & 41.94 & 16.69 & 11.34 \\
    Claude Sonnet 3.7   &   0.0 & 96.9 & 10.6 &  80.0 & 21.6 & 19.6 & \textbf{100.0} &  9.6 &  9.6 &  \textbf{85.0} & 40.1 & 34.6 &  \textbf{5.3} & 36.0 &  3.5 & 58.06 & 37.26 & 16.02 \\
    Claude Sonnet 3.7 T. & 7.1 & 99.9 & 20.9 & \textbf{100.0} & 20.0 & 20.0 & \textbf{100.0} &  9.3 &  9.3 & \textbf{85.0} & 47.8 & 44.7 &  \textbf{5.3} & 36.1 &  4.3 & \textbf{63.44} & 38.98 & 19.92 \\
    Gemini 2 Flash      &   0.0 & 41.8 &  1.4 &  10.0 & 20.5 &  8.8 &  90.0 & 14.5 & 13.9 &   5.0 & 31.0 &  3.2 &  \textbf{5.3} & 34.6 &  2.4 & 23.66 & 27.53 & 6.28 \\
    Gemini 2 Flash T.   & 0.0 & 90.0 &  4.4 &  40.0 & 19.3 & 13.0 & \textbf{100.0} & 15.1 & 15.1 & 25.0 & 55.2 & 21.4 &  0.0 & 30.7 &  3.6 & 35.48 & 39.08 & 12.02 \\
    Llama 405B Instruct &   0.0 & 48.9 &  4.1 &   5.0 & 17.4 &  6.0 &  15.0 &  9.7 &  1.4 &   5.0 & 38.1 &  3.9 &  0.0 & 20.1 &  2.5 & 5.38 & 25.46 & 3.55 \\
    Llama DeepSeek R1   &   7.1 & 54.9 & 12.4 &  90.0 & 18.4 & 17.2 & \textbf{100.0} & 10.1 & 10.1 &  40.0 & 32.1 & 25.4 &  \textbf{5.3} & 17.2 &  3.3 & 51.61 & 24.77 & 13.86 \\
    GPT-o3 mini         &   0.0 & 95.9 &  7.6 & \textbf{100.0} & 18.5 & 18.5 & \textbf{100.0} &  9.7 &  9.7 &  65.0 & 65.0 & 46.7 &  \textbf{5.3} & 34.3 &  6.3 & 58.06 & 41.48 & 18.53 \\
    GPT-o1              &  \textbf{28.6} & 78.8 & 44.6 & \textbf{100.0} & 19.3 & 19.3 & \textbf{100.0} & 10.9 & 10.9 &  70.0 & 32.2 & 27.1 &  \textbf{5.3} &  4.5 &  0.9 & \textbf{63.44} & 26.18 & 19.22 \\
    \hline
    \end{tabular}
    }
\end{table*}

Our experiments reveal a progressive enhancement in LLM-based planners. 
The top-performing models, \textit{Claude Sonnet 3.7 Thinking} and \textit{GPT-o1}, each achieve a success rate of 63.4\%, outperforming their non-reasoning counterparts, \textit{Claude Sonnet 3.7} and \textit{GPT-o3 mini} (Figure \ref{fig:success_rate}). 
This improvement suggests that incorporating reasoning mechanisms enhances planning capabilities.  
A detailed analysis of per-domain performance can be found in Table \ref{tab:results_per_domain}.

However, even the best-performing models do not achieve robust planning. 
Figure \ref{fig:success_timeline} shows that LLMs struggle in more complex domains, raising concerns about whether their success in familiar domains stems from genuine reasoning or simply recalling solutions from pretraining. 
The emergence of reasoning models, such as \textit{Claude Sonnet 3.7 Thinking} and \textit{Gemini 2 Flash Thinking}, demonstrates progress, yet further analysis confirms that LLMs still face challenges in complex domains. 
These findings highlight the necessity of plan validation and execution fidelity in LLM-driven planning for robotic tasks.

\subsection{Plan Length, Executed Actions, and Execution Fidelity}

The plan lengths generated by \textit{Claude Sonnet 3.7}, \textit{Claude Sonnet 3.7 Thinking}, \textit{GPT-o3 mini}, and \textit{Gemini 2 Flash Thinking} align most closely with the baseline established by the \textit{Fast Downward} planner, which produces plans with an average length of 39.56 actions. Among these models, \textit{GPT-o3 mini} is the only one that slightly overestimates plan lengths, generating an average of 41.48 actions per plan. In contrast, the others tend to underestimate plan lengths, with \textit{Claude Sonnet 3.7 Thinking} at 38.98 actions and \textit{Gemini 2 Flash Thinking} at 39.08 actions. \textit{GPT o1} generates notably shorter plans, averaging 26.18 actions, which is largely due to its tendency to conclude that no solution exists in certain cases, leading to no plan generation (Figure \ref{fig:success_timeline}). The shortest average plan length is observed in \textit{Claude Sonnet 3.5}, which generates plans with only 16.69 actions, less than half of the \textit{Fast Downward} baseline.

The executed actions metric evaluates how many actions from a generated plan can be validly executed within the domain’s constraints. The models that perform best according to this metric are \textit{Claude Sonnet 3.7 Thinking} and \textit{GPT-o1}, which execute an average of 19.92 and 19.22 actions, respectively, before violating domain constraints. \textit{GPT-o3 mini} also performs relatively well, executing 18.53 actions on average. In contrast, models with lower plan lengths tend to execute fewer steps. For example, \textit{Claude Sonnet 3.5}, with an average plan length of 16.69 actions, successfully executes only 11.34 steps. The worst-performing model is \textit{Llama 405B Instruct}, which executes just 3.55 actions on average, failing even in the \textit{elevator} domain where all other models perform well.

Execution fidelity, defined as the ratio of executed actions to plan length, provides insight into how well a model adheres to domain constraints. As shown in Figure \ref{fig:execution_fidelity}, \textit{GPT-o1} achieves the highest execution fidelity at 73.4\%, followed by \textit{Llama DeepSeek R1} at 55.9\% and \textit{Claude Sonnet 3.5} at 51.1\%. Notably, despite having the fourth-lowest success rate (41.94\%), \textit{Claude Sonnet 3.5} exhibits relatively high execution fidelity, meaning that while its plans are often valid in terms of execution, their ability to achieve the goal remains uncertain. This contrasts with models such as \textit{Claude Sonnet 3.7 Thinking}, which has a higher success rate (63.44\%) but a lower execution fidelity due to its longer plan length. The lowest execution fidelity is observed in \textit{Llama 405B Instruct}, where only 13.9\% of planned actions can be executed. This highlights the model’s inability to generate valid plans and follow domain constraints effectively.

\subsection{Planning Time}

Planning time varies considerably across models, as shown in Table \ref{tab:planning_time}. The fastest model, \textit{Claude Sonnet 3.5}, requires an average of 14.22 seconds per plan, while the slowest, \textit{Llama DeepSeek R1}, takes 160.15 seconds. Generally, models with explicit reasoning, such as \textit{Claude Sonnet 3.7 Thinking} (112.61 s) and \textit{GPT-o1} (140.51 s), exhibit longer planning times compared to their non-reasoning counterparts. This aligns with the expectation that reasoning increases computational overhead.
There is no strict correlation between planning time and success rate. While \textit{GPT-o1} (140.51 s) achieves the highest success rate (63.44\%), \textit{Llama DeepSeek R1} (160.15 s) has a lower success rate (51.61\%) despite requiring more time. Conversely, some faster models, such as \textit{Claude Sonnet 3.7} (28.90 s), outperform slower ones like \textit{GPT-o3 mini} (98.61 s), indicating that longer planning times do not necessarily result in better plans.
It is important to note that planning time comparisons are not entirely fair, as response times can fluctuate due to varying server loads. Models hosted on heavily used services may experience temporary slowdowns. Additionally, \textit{Llama 405B Instruct} and \textit{Llama DeepSeek R1} are accessed via the \texttt{Fireworks.ai} provider rather than an official hosting service, which may introduce additional latency.

\subsection{Domain-Specific Insights}

The performance of LLM-based planners varies across domains (Figure \ref{fig:success_timeline}). Models perform well in the \textit{blocks} and \textit{elevator} domains, where the problem structures align well with patterns they have likely encountered during training. In contrast, the \textit{barman} and \textit{tidybot} domains remain challenging due to their complex action dependencies and constraints. The consistently high success rates in the \textit{elevator} domain suggest that it may be well-represented in the models' pretraining data. Table \ref{tab:results_per_domain} illustrates how reasoning models improve execution fidelity; however, performance gaps remain in domains that require more elaborate planning strategies.





\section{Discussion, Conclusion \& Future Work}

Robust planning is essential for autonomous robotic systems, where execution failures can lead to critical errors. Our study evaluates whether LLMs can function as general planners, similar to classical solvers like \textit{Fast Downward}, given a domain and problem definition. While models such as \textit{GPT-o1} and \textit{GPT-o3 mini} demonstrate promising success rates and execution fidelity, they still struggle with strict domain constraints, making them unreliable for real-world robotic applications. Execution failures, even partial ones, highlight the difficulty of ensuring action validity over long planning horizons.

Future research should explore hybrid approaches that integrate LLMs with classical planning techniques, combining LLM flexibility with symbolic verification to ensure robustness. Additionally, iterative plan refinement, where an LLM continuously adjusts a plan as execution unfolds, could better align with how humans approach long horizon tasks. If LLMs can truly reason about PDDL structures and maintain internal state consistency, they may serve as adaptive planning components within autonomous robotic systems. While LLM based planning remains immature, its rapid progress suggests potential for new learning driven planning architectures that blend data driven reasoning with structured robotic task execution.

\bibliographystyle{unsrtnat}
\bibliography{mylibclean}

\end{document}